\documentclass[sigconf]{acmart}
\usepackage{comment}
\usepackage{mathtools}
\usepackage{latexsym}
\usepackage{amsmath,amssymb}
\usepackage{graphicx}
\usepackage{xcolor}
\usepackage{url}
\usepackage[linesnumbered,ruled,vlined]{algorithm2e}
\usepackage{balance}
\usepackage{lipsum,flushend}

\newcommand{\myNum}[1]{(\emph{#1})}

\SetCommentSty{mycommfont}

\SetKwInput{KwInput}{Input}                
\SetKwInput{KwOutput}{Output}              

\AtBeginDocument{%
  \providecommand\BibTeX{{%
    \normalfont B\kern-0.5em{\scshape i\kern-0.25em b}\kern-0.8em\TeX}}}

\setcopyright{rightsretained}

\begin{document}
\fancyhead{}

\copyrightyear{2020}
\acmYear{2020}
\acmDOI{01.2345/6789123.4567890}

\acmConference[KDD '20]{Proceedings of the 26th ACM SIGKDD Conference on Knowledge Discovery and Data Mining USB Stick}{August 23--27, 2020}{Virtual Event, USA}
\acmBooktitle{Proceedings of the 26th ACM SIGKDD Conference on Knowledge Discovery and Data Mining USB Stick (KDD '20), August 23--27, 2020, Virtual Event, USA}
\acmDOI{10.1145/3394486.3403231}
\acmISBN{978-1-4503-7998-4/20/08}

\acmSubmissionID{rfp1800}




\title{Unsupervised Paraphrasing via Deep Reinforcement Learning}



\author{A. B. Siddique}
\affiliation{%
  \institution{University of California, Riverside}
}
\email{msidd005@ucr.edu}

\author{Samet Oymak}
\affiliation{%
  \institution{University of California, Riverside}
}
\email{oymak@ece.ucr.edu}

\author{Vagelis Hristidis}
\affiliation{%
  \institution{University of California, Riverside}
}
\email{vagelis@cs.ucr.edu}

\renewcommand{\shortauthors}{Anonymous, et al.}

\newcommand{\reminder}[1]{ {\color{red}[[[ { \bf #1 } ]]] }}

\begin{abstract}
Paraphrasing is expressing the meaning of an input sentence in different wording while maintaining fluency (i.e.,~grammatical and syntactical correctness).
Most existing work on paraphrasing use supervised models that are limited to specific domains (e.g., image captions). Such models can neither be straightforwardly transferred to other domains nor generalize well, and creating labeled training data for new domains is expensive and laborious. The need for paraphrasing across different domains and the scarcity of labeled training data in many such domains call for exploring unsupervised paraphrase generation methods. 
We propose Progressive Unsupervised Paraphrasing~(PUP): a novel unsupervised paraphrase generation method based on deep reinforcement learning (DRL).
PUP uses a variational autoencoder (trained using a non-parallel corpus) to generate a seed paraphrase that warm-starts the DRL model. Then, PUP progressively tunes the seed paraphrase guided by our novel reward function which combines semantic adequacy, language fluency, and expression diversity measures to quantify the quality of the generated paraphrases in each iteration without needing parallel sentences. 
Our extensive experimental evaluation shows that PUP outperforms unsupervised state-of-the-art paraphrasing techniques in terms of both automatic metrics and user studies on four real datasets. We also show that PUP outperforms domain-adapted supervised algorithms on several datasets. Our evaluation also shows that PUP achieves a great trade-off between semantic similarity and diversity of expression.

\end{abstract}




\maketitle

\section{Introduction}
\label{intro}
Paraphrasing is the task of generating a fluent output sentence, given an input sentence, to convey the same meaning in different wording. It is an important problem in Natural Language Processing (NLP) with a wide range of applications such as summarization~\cite{kissner2006summarizing}, information retrieval~\cite{knight2000statistics}, question answering~\cite{mckeown1983paraphrasing}, and conversational agents~\cite{shah2018building}. Most of the previous paraphrasing work~\cite{prakash2016neural,li2017paraphrase,gupta2018deep} has focused on \textit{supervised} paraphrasing methods, which require large corpora of parallel sentences (i.e., input and corresponding paraphrased sentences) for training. Unlike large datasets in neural machine translation, there are not many parallel corpora for paraphrasing, and they are often domain-specific, e.g., Quora is a questions dataset, and MSCOCO is an image captioning dataset.
Acquiring big parallel datasets for paraphrasing across many domains is not scalable because it is expensive and laborious. Moreover, a model trained in one domain does not generalize well to other domains~\cite{li2019decomposable}. 

The abundance of domains and applications that could benefit from paraphrasing calls for exploring unsupervised paraphrasing, which is still in its infancy. There are relatively few works on unsupervised paraphrasing such as Variational Autoencoder (VAE)~\cite{bowman2015generating}, Constrained Sentence Generation by Metropolis-Hastings Sampling (CGMH)~\cite{miao2019cgmh}, and Unsupervised Paraphrasing by Simulated Annealing (UPSA)~\cite{liu2019unsupervised}. 
Although unsupervised approaches have shown promising results, the probabilistic sampling based approaches such as VAE~\cite{bowman2015generating} and CGMH~\cite{miao2019cgmh} are less constrained, and they produce paraphrases that lack semantic similarity to the input. On the other hand, 
UPSA~\cite{liu2019unsupervised} does not effectively explore the entire sentence space, resulting in paraphrases that are not different enough from the input.

Given the success of Deep Reinforcement Learning (DRL)~\cite{sutton1998introduction} in a wide range of applications such as Atari games~\cite{mnih2013playing}, alphaZero~\cite{silver2017mastering}, and supervised paraphrasing~\cite{li2017paraphrase}, can DRL also help boost the performance of unsupervised paraphrase generation?
To the best of our knowledge, this is the first work to employ DRL in unsupervised paraphrase generation, which is challenging due to the following reasons:
\myNum{i} DRL is known to not work well with large vocabulary sizes when starting with a random policy (i.e., random exploration strategy)~\cite{dayan2008reinforcement,li2017paraphrase}; 
\myNum{ii} paraphrasing is a multi-step (word-by-word) prediction task, where a small error at an early time-step may lead to poor predictions for the rest of the sentence, as the error is compounded over the next token predictions; and 
\myNum{iii} it is challenging to define a reward function that incorporates all the characteristics of a good paraphrase with no access to parallel sentences (i.e., the unsupervised setting).

\begin{figure*}[t!]
  \centering
  \includegraphics[width=\linewidth]{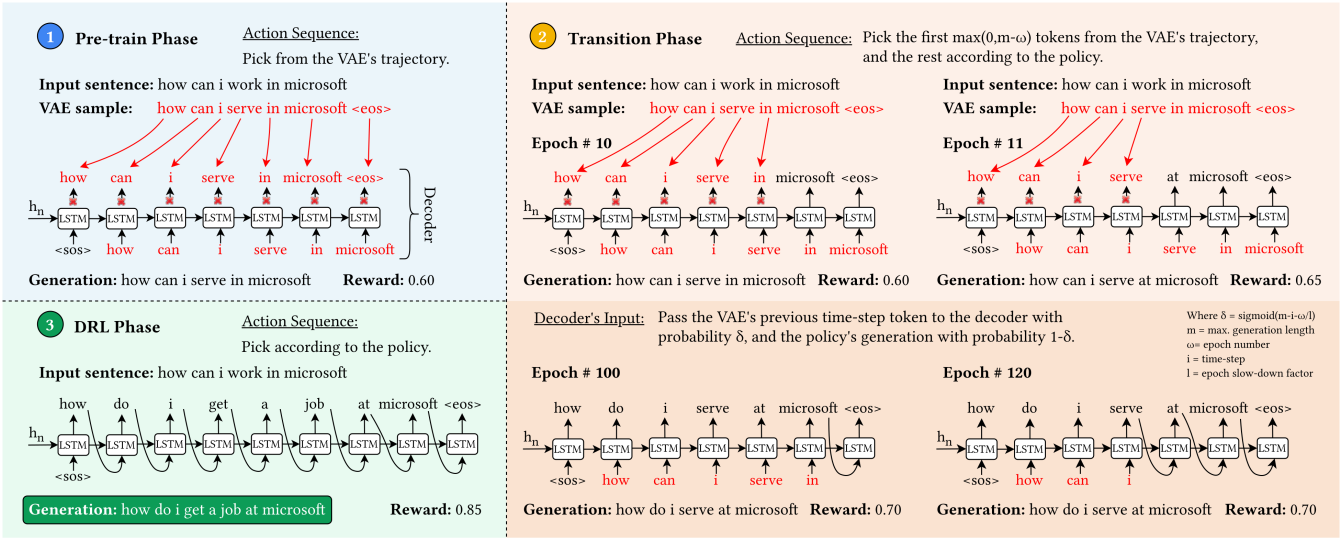}
  \caption{Illustration of the decoding process of the proposed unsupervised paraphrasing method: PUP. Red and black color tokens represent the output from VAE and the DRL's chosen action sequences respectively. Whereas the sentence in green is the final paraphrased sentence generated by PUP for the given input sentence.}
  \label{fig:example}
  \vspace{-0.25cm}
\end{figure*}

Our proposed method, \textit{Progressive Unsupervised Paraphrasing (PUP)}, progressively trains a DRL-based model for unsupervised paraphrasing and addresses the aforementioned three challenges using the following techniques:

\noindent $\bullet~~$ {\bf{Unsupervised warm-start of DRL:}} PUP warm-starts reinforcement learning by an unsupervised pre-trained VAE~\cite{bowman2015generating}, which acts as an expert~\cite{daume2009search,ross2011reduction} in the pre-training phase. The pre-trained VAE saves the DRL model from expensive global exploration during the initial training phase. Remarkably, the proposed technique is the first instance that can successfully warm-start DRL with an unsupervised model. At the end of DRL training, our DRL model achieves up to $54$\% higher reward compared to the initial VAE model.
We expect that our idea of warm-starting DRL models in an unsupervised fashion may have implications on a broader range of NLP problems with limited labels. \newline
\noindent $\bullet~~${\bf{Progressive transition for seq2seq DRL:}} Another major issue DRL models face is the accumulation of error over the predictions of future tokens. This is particularly significant during the initial exploration of the space. To overcome this, we use a progressive transition that takes advantage of the Sequence-to-Sequence (seq2seq)~\cite{sutskever2014sequence} nature of the problem by transitioning between algorithms (e.g., VAE to DRL) token by token, as shown in Figure~\ref{fig:example}. Instead of taking actions according to the initial policy (i.e., random action), the model chooses VAE's output as the action, and then incrementally (i.e., one token per epoch) allows the agent to take actions according to the DRL policy. This technique greatly facilitates the convergence of DRL to models with high rewards and is at the heart of the success of DRL.



\noindent$\bullet~~$ {\bf{Unsupervised reward function for paraphrasing:}} We propose a novel reward function for the DRL model that can measure the quality of the generated paraphrases when no parallel sentences are available. This is accomplished by incorporating the most desirable qualities of a good paraphrase , informed on the paraphrasing literature~\cite{zhao2010paraphrases,zhao2009application,zhao2010leveraging,chen2011collecting,metzler2011empirical,sun2012joint}. Our reward function is a combination of semantic adequacy, language fluency, and diversity in expression. 

Figure~\ref{fig:example} provides an illustration of the decoding process of PUP.
First, the decoder of the DRL model relies on the VAE's sample to pick its actions in the pre-train phase.
Then, in the transition phase, the model gradually starts taking actions according to its policy.
Finally, in the DRL phase, the model picks actions entirely according to its policy to maximize the expected reward.
For example, when our DRL model is pre-trained with the VAE sample  \textit{"how can i serve in microsoft"}, our fully-trained DRL model amazingly generates the paraphrase \textit{"how do i get a job at microsoft"}.
We evaluate PUP on four real datasets and compare it against state-of-the-art unsupervised paraphrasing techniques; we show that PUP outperforms them in all standard metrics.
We also conduct a human study, which demonstrates that human evaluators find PUP's paraphrases to be of higher quality compared to other methods' paraphrases across several carefully selected measures.
Moreover, we consider comparisons against domain-adapted models -- i.e., models trained on one dataset such as Quora in a supervised setting and then domain-adapted for another dataset such WikiAnswers in an unsupervised fashion. 
Remarkably, PUP outperforms domain-adapted supervised paraphrasing methods in datasets where applicable.

The rest of the paper is organized as follows. Background is discussed in Section~\ref{background}, and an overview of PUP is presented in Section~\ref{overview}. The details of PUP are described in Section~\ref{drl}. Sections~\ref{experiments} and~\ref{results} present the experimental setup and results, respectively. Section~\ref{related} presents the related work, and Section~\ref{conclusion} concludes the paper. 
\section{Background} 
\label{background}




\subsection{Encoder-Decoder Framework}
\label{seq2seq}
An encoder-decoder model (e.g.,~seq2seq) strives to generate a target sequence (i.e., paraphrase) $Y = (y_1, y_2, \cdots, y_m)$ given an input sequence $X = (x_1, x_2, \cdots , x_n)$, where $m$ and $n$ are target and input sequence lengths respectively.
First, the encoder transforms the input sequence $X$ into a sequence of hidden states $(h_1,h_2, \cdots, h_n)$ employing RNN units such as Long Short-Term Memory (LSTM)~\cite{hochreiter1997lstm}.
The encoder reads the input sequence, one token at a time, until the end of the input sequence token occurs and converts it to hidden state $h_i = Encoder(h_{i-1}, emb(x_i))$ by considering the word embedding of the input token $x_i$ and the previous hidden state $h_{i-1}$ at time-step $i$. $Encoder(.)$ is a non-linear mapping function and $emb(.)$ maps the given word into a high dimensional space. The decoder utilizes another RNN to generate the paraphrased (i.e., target) sequence $Y$. The decoder is initialized with the last hidden state $h_n$, and generates one token at a time, until the end of sentence token (i.e., $<eos>$) is generated.
At time-step $i$, the generation is conditioned on the previously generated words $\hat y_{i-1}, \cdots, \hat y_1$ and the current decoder hidden state $h^\prime_i$:
\begin{equation}
  P(y_i | \hat y_{i-1}, \cdots, \hat y_1, X)= softmax(Decoder(h^\prime_i, y_{i-1})).
  \label{eq:enc-dec}
\end{equation}

Where $Decoder(.)$ is a non-linear mapping function and $softmax(.)$ converts the given vector into a  probability distribution. Such an encoder-decoder model is typically trained by minimizing the negative log-likelihood of the input-target pairs.
However, since we do not have access to target sentences in the unsupervised paraphrase generation task, we utilize the reinforcement learning framework.
 
\subsection{VAE: Variational Autoencoder}
\label{vae}

VAE~\cite{kingma2013auto,rezende2014stochastic} 
is a deep generative model for learning a nonlinear latent representation $z$ from data points $X$. It is trained in an unsupervised fashion for the following loss function:
\begin{equation}
\label{eq:vae_loss}
\begin{split}
 \mathcal{L}_{vae}(\varphi, \phi)= - {\mathbb{E}}_{q_\varphi(z | X)}\big[\log p_\phi (X | z) \big]  + {\mathbb{K}}{\mathbb{L}}(q_\varphi(z|X)||p(z)), 
\end{split}
\end{equation}
where $q_\varphi(z | X)$ is the encoder with parameters $\varphi$ that encodes the data points $X$ into a stochastic latent representation $z$; $p_\phi (X | z)$ is the decoder with the parameters $\phi$ that strives to generate an observation $X$ given the random latent code $z$; and $p(z)$ is prior distribution, i.e., standard normal distribution $\mathcal{N}(0,{\mathrm{I}})$.
The first term in Equation~\ref{eq:vae_loss} is the negative log-likelihood loss for the reconstruction of the data points $X$.
The second term is used to measure Kullback-Leibler (${\mathbb{K}}{\mathbb{L}}$) divergence between the the encoder’s distribution $q_\varphi(z | X)$ and the prior distribution $p(z)$.
At inference time, sentences are sampled~\cite{bowman2015generating} from the learned latent representation $z$.
In this work, VAE is employed to provide a warm-start to the DRL-based paraphrasing model so that it does not start from a random policy.
\section{Overview of PUP}
\label{overview}

This section provides an overview of the progressive training phases of PUP (Figure~\ref{fig:example}). It consists of three phases: pre-train, progressive transition, and DRL. \newline
\textit{Pre-train phase:}
For tasks like unsupervised paraphrasing, the big vocabulary impedes the learning process of DRL models. It becomes practically infeasible to train such a model based on the reward alone.
To address this issue, we employ a pre-trained VAE (trained on a non-parallel corpus) to provide a warm-start to the DRL model. That is, the output of VAE is used to pick action sequences instead of the agent policy's output.
We can think of it as demonstrating the expert's (VAE) actions to DRL, where the expert is an unsupervised model. \newline
\textit{Progressive transition phase:}
The next critical step is to gracefully transition from following the expert's actions to taking actions according to the policy (i.e., DRL decoder's distribution). An abrupt transition can obstruct the learning process due to the nature of the task, i.e., multi-step prediction, where error accumulates. Especially, an inappropriate sample at an early stage of the sentence (i.e., first few words) may lead to a poor eventual paraphrase generation (i.e., ungrammatical or semantically unfaithful).
We propose an intuitive way to pick the first $max(0,m-\omega)$ tokens from VAE's output, and pick the rest according to the agent policy, where $m$ is the length of the generated sentence and $\omega$ is the epoch number.
Moreover, we pass the output of VAE to the decoder's next time-step with a decreasing probability $\delta$ (i.e., decreasing with respect to $\omega$), and the DRL's generation otherwise. This helps with mitigating the accumulation of error, especially in the beginning of the transition phase when the model is expected to make mistakes. 
\newline
\textit{DRL phase:}
Finally, the model is trained to produce an optimized policy by sampling sentences according to its policy and maximizing its expected reward, which is a combination of the semantic adequacy, language fluency, and diversity in expression. 

\begin{figure}[t!]
  \centering
  \includegraphics[width=\linewidth]{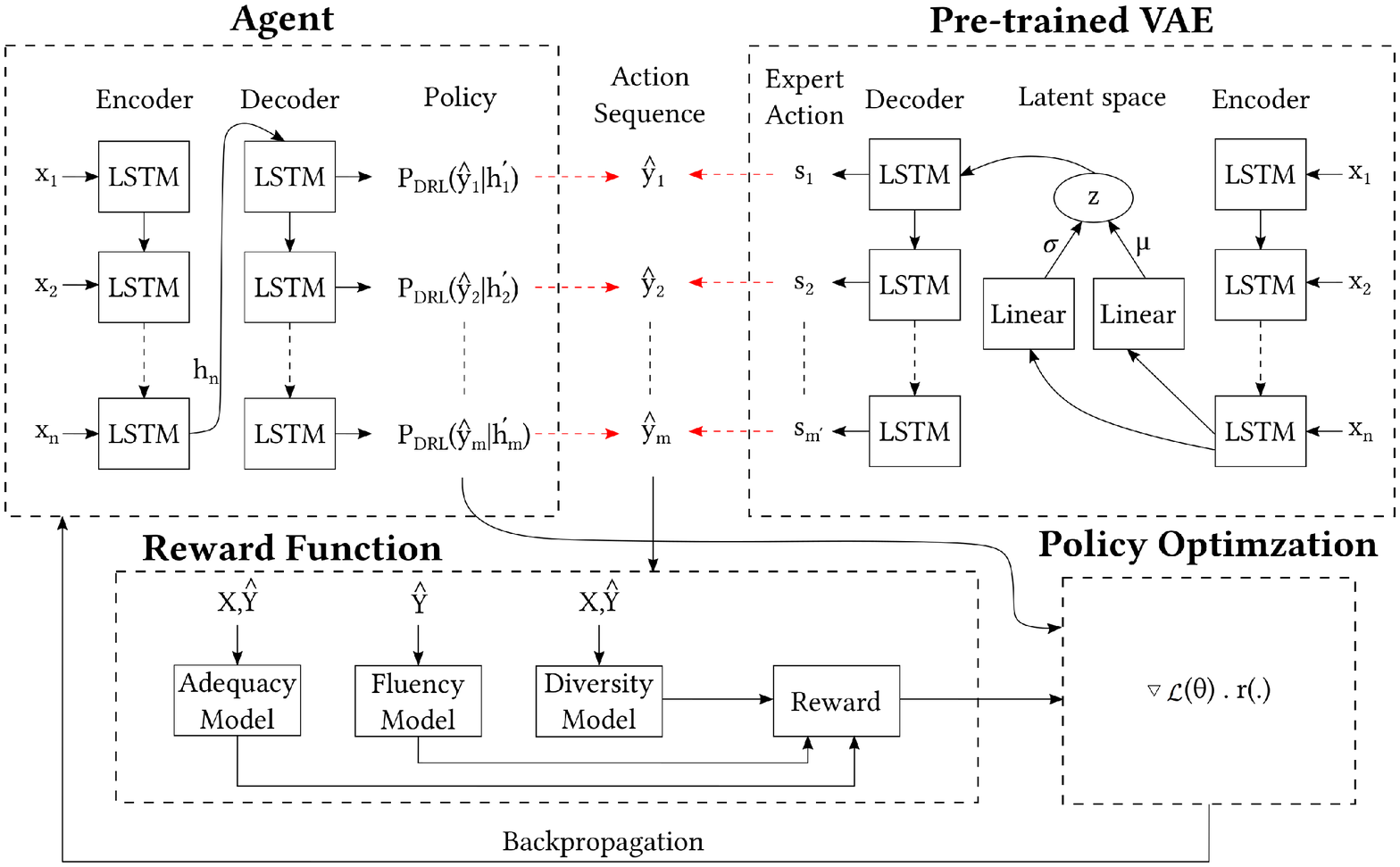}
  \caption{Deep reinforcement learning paradigm for unsupervised paraphrase generation.}
  \label{fig:drl_paradigm}
  \vspace{-0.2cm}
\end{figure}

Figure~\ref{fig:drl_paradigm} presents an overview of the DRL paradigm, where action sequences are picked either from VAE's output or the agent policy ( highlighted by red dashed arrows) depending on the different phases. 


\section{Progressive Unsupervised Paraphrasing (PUP)}
\label{drl}

We first describe how to incorporate DRL for the unsupervised paraphrasing task, then the proposed reward function, and finally we describe the details of PUP.

\subsection{Reinforcement Learning Paradigm}
\label{rl}
The reinforcement learning paradigm for unsupervised paraphrasing is presented in Figure~\ref{fig:drl_paradigm}. 
In DRL terminology, the encoder-decoder model (Section~\ref{seq2seq}) acts as an agent, which first encodes the input sentence $X$ and then generates the paraphrased version $\hat Y$. 
At time-step $i$, the agent takes an action $\hat y_i \in V$ according to the policy $P_{DRL}(\hat y_i | \hat y_{1:i-1}, X)$ (see Equation~\ref{eq:enc-dec}), where $V$ represents the possible action space (i.e., vocabulary for generation).
The hidden states of the encoder and the previous outputs of the decoder constitute the state. 
The agent (i.e., model) keeps generating one token at a time, until the end of sentence token (i.e., $<eos>$) is produced, which completes the action sequence (i.e., trajectory) $\hat Y = (\hat y_1, \hat y_2, \cdots, \hat y_m)$. The policy is optimized by maximizing the expected reward $r$ for the action sequences.

\subsection{Paraphrasing Reward}
\label{para-reward}
Automatic quality measures for machine translation (or paraphrasing) such as BLEU~\cite{papineni2002bleu}, Rouge~\cite{hovy2006automated}, TER~\cite{snover2006study}, and METEOR~\cite{banerjee2005meteor} only work when parallel sentences (i.e., targets or references) are available. 
We propose a novel reward function that incorporates all the characteristics of a good paraphrase and does not require  parallel sentences. The most desired qualities of a good paraphrase~\cite{zhao2010paraphrases,zhao2009application,zhao2010leveraging,chen2011collecting,metzler2011empirical,sun2012joint} include: semantic adequacy (i.e., similarity in meaning), language fluency (i.e., grammatical correctness), and diversity of expression (i.e., sentence dissimilarity). We define the reward $r(X, \hat Y)$ of an output sequence $\hat Y$ generated by the DRL model for input $X$ as a combination of the above components: 
\begin{equation}
  r(X, \hat Y)=  \alpha \ldotp r_{Sim}(X, \hat Y) + \beta \ldotp r_{F}(\hat Y) + \gamma \ldotp r_{D}(X, \hat Y),
\end{equation}
\noindent where $r_{Sim}(X, \hat Y)$, $r_{F}(\hat Y)$ and, $r_{D}(X, \hat Y)$  $\in [0,1]$.
$r_{Sim}(X, \hat Y)$ is the semantic similarity between input $X$ and generated paraphrase $\hat Y$.
$r_{F}(\hat Y)$ captures whether the generated sentence $\hat Y$ is grammatically correct or not.
$r_{D}(X, \hat Y)$ measures the diversity between $X$ and $\hat Y$.
$\alpha$, $\beta$, and $\gamma$ $\in [0,1]$ are respective weights. Each component is described below.

\noindent{\bf{Semantic Adequacy:}} The semantic adequacy reward $r_{Sim}(X, \hat Y)$ makes sure that the generated paraphrase $\hat Y$ is similar in meaning to the input sequence $X$.
We use the universal sentence encoder~\cite{cer2018universal}, as it has achieved state-of-art results for semantic textual similarity on the STS Benchmark~\cite{cer2017semeval} and it provides a straightforward process to incorporate it in any implementation.
In a nutshell, it is trained with a deep averaging network (DAN) encoder, and it generates $512$-dimension embedding vector for arbitrary length sentence(s). Then, the semantic similarity can be calculated using the cosine similarity of the vectors $v_X$ and $v_{\hat Y}$, which are embedding vectors for the input sequence $X$ and the paraphrased sequence $\hat Y$, respectively. 
\begin{equation}
  r_{Sim}(X, \hat Y) = \cos (v_X, v_{\hat Y}) = \frac{v_X \ldotp v_{\hat Y} } { {\| v_X \|} {\| v_{\hat Y} \|}}
\end{equation}

\noindent{\bf{Language Fluency:}}
The fluency reward $r_{F}(\hat Y)$ measures the grammatical correctness of the generated paraphrase $\hat Y$. Since language models such as n-grams~\cite{heafield2011kenlm} and neural models~\cite{bengio2008neural} are trained to predict the next token given previous tokens, they can be used to score sentences for fluency. Recently, the Corpus of Linguistic Acceptability (CoLA)~\cite{warstadt2019neural} has produced the state-of-art results on the grammatical acceptability for in-domain as well as out-of-domain test sets. In its simplest form, CoLA~\cite{warstadt2019neural} utilizes ELMo-Style (Embeddings from Language Models) and pooling classifier, and it is trained in a supervised fashion. We use a pre-trained CoLA~\cite{warstadt2019neural} to score our generated paraphrased sequences $\hat Y$.

\noindent{\bf{Expression Diversity:}}
The expression diversity reward $r_{D}(X, \hat Y)$ encourages the model to generate tokens that are not in the input sequence $X$. One of the simplest methods to measure the diversity, inverse Jaccard similarity (i.e., $1 - \text{Jaccard Similarity}$), could be used.
In this work, we use 
n-grams dissimilarity. To measure the diversity in expression, we use the inverse BLEU of input sequence $X$ and the generated sequence $\hat Y$, which is computed using $1$ - BLEU( $X$, $\hat Y$).
The average of the uni-gram and bi-gram inverse BLEU scores are used in $r_{D}(X, \hat Y)$. 


\noindent{\bf{Combining the three components:}} 
In practice, a reward function that can force the DRL model to generate good quality paraphrases must maintain a good balance across the reward components (i.e., semantic similarity, fluency, and diversity).
For example, generating diverse words at the expense of losing too much on the semantic adequacy or fluency is not desirable. Similarly, copying the input sentence as-is to the generation is clearly not a paraphrase (i.e., cosine similarity = $1$). To achieve this, we impose strict criteria on the components of the reward function as given below: 
\begin{flalign}
\label{eq:condition_sim}
& \: r_{Sim}(X, \hat Y)= \left\{
  \begin{array}{@{}ll@{}}
    r_{Sim}(X, \hat Y), & \text{if}\ \tau_{min} \leq r_{Sim}(X, \hat Y) \leq \tau_{max}  \\
    0, & \text{otherwise}
  \end{array}\right.
\end{flalign} 

\begin{flalign}
\label{eq:condition_fluency}
& \; \quad  r_{F}(\hat Y)= \left\{
  \begin{array}{@{}ll@{}}
    r_{F}(\hat Y), & \text{if}\ r_{F}(\hat Y)\geq \lambda_{min} \\
    0, & \text{otherwise}
  \end{array}\right. &
\end{flalign} 

\begin{flalign}
\label{eq:condition_diversity}
& \; r_{D}(X, \hat Y)= \left\{
  \begin{array}{@{}ll@{}}
    r_{D}(X, \hat Y), & \text{if}\ r_{Sim}(X, \hat Y)\geq \tau_{min}, r_{F}(\hat Y)\geq \lambda_{min}  \\
    0, & \text{otherwise}
  \end{array}\right.
\end{flalign} 
Equation~\ref{eq:condition_sim} makes sure that the model does not copy the input sentence as-is to the generation (i.e., condition: $r_{Sim}(X, \hat Y) \leq \tau_{max}$) to enforce the diversity in expression, and does not generate random sentence, which has very low similarity with the input (i.e., condition: $ r_{Sim}(X, \hat Y) > \tau_{min}$). Equation~\ref{eq:condition_fluency} penalizes the generations that are not fluent. Finally, diverse words (i.e., Equation~\ref{eq:condition_diversity}) get rewarded only if the generated sentence achieves a reasonable score on the semantic similarity (i.e., condition: $r_{Sim}(X, \hat Y)\geq \tau_{min}$) and fluency (i.e., condition: $r_{F}(\hat Y)\geq \lambda_{min}$).
Note that a diversely expressed output sentence, which is not fluent or is not close in meaning to the input sentence needs penalization so that the model may learn a policy that generates not only diverse sentences but also fluent and semantically similar to the input.
The objective of combining all the constraints is to ensure competitive outputs in all metrics and to penalize the model for poor generations on any metric.
The weights for each component in the reward (i.e., $\alpha$, $\beta$, and $\gamma$), and thresholds (i.e., $\tau_{min}$, $\tau_{max}$, and $\lambda_{min}$) for  Equations~\ref{eq:condition_sim},~\ref{eq:condition_fluency}, and ~\ref{eq:condition_diversity} can be defined based the application needs.



\subsection{Progressively Training the DRL}
\label{training}

The training algorithm optimizes the policy (i.e., encoder-decoder model's distribution $P_{DRL}(. | X)$) to maximize the expected reward $r(.)$ for the generated action sequence $\hat Y = (\hat y_1, \hat y_2, \cdots, \hat y_m)$.
The loss for a single sample from the possible action sequences is:
\begin{equation}
  \mathcal{L} (\theta) = - {\mathbb{E}}_{(\hat y_1, \hat y_2, \cdots, \hat y_m)}  \sim  P_{DRL}(. | X) [r (\hat y_1, \hat y_2, \cdots, \hat y_m)].
\end{equation}
The loss is the negative expected reward for the action sequences. Infinite number of possible samples make the expectation calculations infeasible, thus it is approximated~\cite{williams1992simple}. The gradient for the $\mathcal{L} (\theta)$ is:
\begin{equation}
 \nabla \mathcal{L} (\theta) \approx \sum_{i=1}^{m} \nabla \log P_{DRL}(\hat y_i | \hat y_{1:i-1}, X) [r (\hat y_1, \hat y_2, \cdots, \hat y_m)].
\end{equation}

The training process for the DRL-based unsupervised paraphrase generation model is outlined in Algorithm~\ref{rl_algo}. We explain each of the training phases below. 
Note that the pre-trained VAE and the DRL model share the same vocabulary.

\begin{algorithm}[t!]
\DontPrintSemicolon
  
  \KwInput{A non-parallel training example $X = (x_1,x_2, \cdots, x_n)$, a paraphrase generated by VAE $S = (s_1,s_2, \cdots, s_{m \prime})$, probability $\delta$ to pass VAE's output as input to decoder, probability $\epsilon$ to sample according to the policy, epoch number $\omega$, pre-training status $\rho$, and the learning rate $\eta$.}
 Initialize $\mathcal{L} (\theta) \gets 0$ \;
  \For{i=1,$\cdots$, m}
    {
    $vae\_in \gets Uniform(0,1)$ \newline
    \If  {$vae\_in < \delta$}
    {
    $\hat y_{i-1} \gets s_{i-1}$
    }
        \If{$i \leq m - \omega$ OR $\rho = True$}
        {
        $\hat y_i \gets s_i$
        }
        \Else
        {
        $explore \gets Uniform(0,1)$ \newline
        \If{$explore < \epsilon$}
        {
        $\hat y_i \gets$ Sample $P_{DRL}(\hat y_i | h^\prime_i, \hat y_{i-1})$
        }
        \Else
        {
        $\hat y_i \gets$ Argmax $P_{DRL}(\hat y_i | h^\prime_i, \hat y_{i-1})$
        }
        }
    $\mathcal{L} (\theta) \gets  \mathcal{L} (\theta) + \log P_{DRL}(\hat y_i | h^\prime_i, \hat y_{i-1})$
    }
  $\theta \gets$ $\theta$ + $\eta$ .  ($\nabla \mathcal{L} (\theta) $ . $r(X, \hat Y))$
    
\caption{Progressively training DRL-based method.}
\label{rl_algo}
\end{algorithm}

\noindent{\bf{Pre-train Phase:}}
Pre-training is a critical step for DRL to work in practice. Since one of the main contributions of this work is to make DRL work in purely unsupervised fashion for the task of paraphrase generation, the pre-training step also has to be unsupervised. We use VAE~\cite{bowman2015generating}, which is trained in an unsupervised way, and serves as a decent baseline in unsupervised paraphrase generation tasks~\cite{miao2019cgmh}.
The pre-trained VAE (section~\ref{vae}) guides as an expert in the pre-train phase to provide a warm-start. Line 6 in Algorithm~\ref{rl_algo} refers to the pre-train phase. At time-step $i$, the algorithm picks VAE's sample $s_i$ as the action $\hat y_i$. 
The loss $\mathcal{L} (\theta)$ is computed and accumulated (see line 12 in Algorithm~\ref{rl_algo}). Once, the action sequence is complete (i.e., $(\hat y_1, \hat y_2, \cdots, \hat y_m)$), the reward $r$ is calculated and parameters $\theta$ are updated (line 13). This step is a requisite for the DRL model to work in practice for unsupervised paraphrasing.

\noindent{\bf{Transition Phase:}}
Once the model is able to generate sensible sentences, the next critical step is to progressively allow the agent (i.e., encoder-decoder model) to take actions according to its policy.
Line 5 in Algorithm~\ref{rl_algo} refers to whether to take action according to the policy $P_{DRL}$ or to utilize VAE's output $S$.
First $max(0, m - \omega)$ tokens are picked from VAE, and the rest are sampled according to the policy $P_{DRL}(\hat y_i | h^\prime_i, \hat y_{i-1})$ at time-step $i$, where $m$ is the length of the generation (i.e., action sequence) and $\omega$ is the epoch number.
This way, the model picks all tokens from VAE in epoch $0$, and in epoch $1$, the model is allowed to pick only the last token according to its policy, and so on.
Similarly, by epoch $m$, the model learns to pick all the tokens according to its policy and none from the VAE.
The intuition behind this gradual token-by-token transition is that mistakes at earlier tokens (i.e., words at the beginning of the sentence) can be catastrophic, and picking the last few tokens is relatively easy.
Moreover, allowing the model to pick according to its policy as soon as possible is also needed, hence we employ gradual transitioning.

Since we allow the DRL model to pick according to its policy at an early stage in the transition phase, the model is expected to make mistakes. However, letting these errors compound over the next predictions may result in never being able to generate sufficiently good samples that can get high rewards. Lines 3-4 in Algorithm~\ref{rl_algo} attempt to overcome this issue by passing the VAE's previous token $S_{i-1}$ to the decoder as input at time-step $i$ with probability $\delta =sigmoid(m-i -\omega/l) ) \in [0,1]$, where $m$ is the length of the output sentence, $\omega$ is the epoch number, and $l$ is the slow-down factor to decay the probability $\delta$ as $\omega$ grows.
It is similar to the above gradual transitioning, but $l$ times slower and probabilistic. The intuition behind the slow progressive transition is that if the DRL model samples wrong token, passing the VAE's output to upcoming time-step's decoder would eliminate the accumulation of error in the beginning of the transition phase. 



\noindent{\bf{DRL Phase:}}
The DRL phase is the classic reinforcement learning, where the agent takes action $\hat Y$ according to its policy $P_{DRL}$, gets reward $r$, and optimizes its policy to maximize its expected reward.
Greedy decoding impedes the exploration of the space, whereas continuous exploring is also not a desirable behaviour. To keep a balance between exploration (i.e., sample) and exploitation (i.e., argmax), we use a probabilistic decaying mechanism for exploration with probability $\epsilon = \kappa^{\omega}$, where $\kappa \in [0,1]$ is the constant to control the decay rate of the probability $\epsilon$ as $\omega$ grows. Lines 7-11 in Algorithm~\ref{rl_algo} refer to this phase.  Pre-trained VAE is used as a baseline model in this phase.
\section{Experimental Setup}
\label{experiments}

\begin{table}
  \caption{Statistics about paraphrase datasets}
  \label{tab:datasets}
  \begin{tabular}{lrrrr}
    \toprule
    Dataset&Train&Valid&Test&Vocabulary\\
    \midrule
    Quora &	$117$K	&$3$K	&$20$K	&$8$K\\
    WikiAnswers&	$500$K&	$6$K&	$20$K&	$8$K\\
    MSCOCO &	$110$K&	$10$K&	$40$K&	$10$k\\
    Twitter &	$10$K&	$2$K&	$2$K&	$8$K \\
  \bottomrule
\end{tabular}
\end{table}

\begin{table*}[t!]
\caption{Performance of the unsupervised and domain-adapted methods on Quora and WikiAnswers datasets.}
\label{tab:quora}
\begin{tabular}{llllllllll}
\toprule
                                             &                   & \multicolumn{4}{c}{Quora}        & \multicolumn{4}{c}{WikiAnswers} \\ \cline{3-10} 
                                             & Method            & i-BLEU & BLEU  & Rouge1 & Rouge2 & i-BLEU & BLEU  & Rouge1 & Rouge2 \\ \midrule
{Supervised + } & Pointer-generator & $5.04$   & $6.96$  & $41.89$  & $12.77$  & $21.87$  & $27.94$ & $53.99$  & $20.85$  \\
        domain adapted     & Transformer+Copy  & $6.17$   & $8.15$  & $44.89$  & $14.79$  & $23.25$  & $29.22$ & $53.33$  & $21.02$  \\
                                             & Shallow fusion    & $6.04$   & $7.95$  & $44.87$  & $14.79$  & $22.57$  & $29.76$ & $53.54$  & $20.68$  \\
                                             & MTL               & $4.90$   & $6.37$  & $37.64$  & $11.83$  & $18.34$  & $23.65$ & $48.19$  & $17.53$  \\
                                             & MTL+Copy          & $7.22$   & $9.83$  & $47.08$  & $19.03$  & $21.87$  & $30.78$ & $54.10$   & $21.08$  \\
                                             & DNPG              & $10.39$  & $16.98$ & $56.01$  & $28.61$  & $\textbf{25.60}$   & $35.12$ & $56.17$  & $23.65$  \\ \midrule
{Unsupervised} & VAE &	$8.16$	&$13.96$&	$44.55$ &	$22.64$ &$17.92$	& $24.13$	& $31.87$ &	$12.08$ \\               
                                         & CGMH              & $9.94$   & $15.73$ & $48.73$  & $26.12$  & $20.05$  & $26.45$ & $43.31$  & $16.53$  \\
                                             & UPSA              & $12.02$  & $18.18$ & $56.51$  & $\textbf{\underline{30.69}}$  & $24.84$  & $32.39$ & $54.12$  & $21.45$  \\
                                             & PUP               & $\textbf{\underline{14.91}}$  & $\textbf{\underline{19.68}}$ & $\textbf{\underline{59.77}}$  & $30.47$  & $\underline{25.20}$   & $\textbf{\underline{38.22}}$ & $\textbf{\underline{58.88}}$  & $\textbf{\underline{26.72}}$  \\ 
\bottomrule
\end{tabular}
\end{table*}

In this section, we describe the datasets, competing approaches, evaluation metrics, and the implementation details of PUP.

\subsection{Dataset}
\label{datasets}
We use Quora~\cite{QuoraQue91:online}, 
WikiAnswers~\cite{fader2013paraphrase}, MSCOCO~\cite{lin2014microsoft}, and Twitter~\cite{lan2017continuously} datasets to evaluate the quality of the paraphrase generated by PUP and other competing approaches. Table~\ref{tab:datasets} presents key statistics about the datasets.
It is important to mention that although these datasets have parallel sentences, we don't use them for training nor for validation.
We only use parallel sentences to compute the evaluation results on the respective testing sets.


\noindent {\bf{Quora}} is a popular dataset for duplicate question detection annotated by humans which has been used for evaluating paraphrase quality as well, since a pair of duplicate questions can also be considered paraphrases of each other. We follow the training, validation, and testing splits used by~\cite{miao2019cgmh,liu2019unsupervised} for a fair comparison. \newline
\noindent {\bf{WikiAnswers}} 
contains $2$M duplicate question-paraphrase pairs. 
We use $500K$ non-parallel sentences for training, following previous works~\cite{li2019decomposable,liu2019unsupervised}. \newline
\noindent {\bf{MSCOCO}} is an image captioning dataset that has over $120$K images, each captioned by $5$ different human annotators.
Since all the captions for an image can be thought of as paraphrases, it has also been utilized for the paraphrasing task. We follow the standard splitting~\cite{lin2014microsoft} and evaluation protocols~\cite{liu2019unsupervised,prakash2016neural} in our experiments. \newline
\noindent {\bf{Twitter}} dataset is also annotated by humans for duplicate detection. We use the standard train/test split~\cite{lan2017continuously}, and further split the training set to create a validation set (i.e., $2$K sentences).

\begin{table*}[t!]
\caption{Performance of Unsupervised approaches on MSCOCO and Twitter dataset.}
\label{tab:mscoco}
\begin{tabular}{lllllllll}
\toprule
       & \multicolumn{4}{c}{MSCOCO}       & \multicolumn{4}{c}{Twitter}      \\
       \cline{2-9} 
Method & i-BLEU & BLEU  & Rouge1 & Rouge2 & i-BLEU & BLEU  & Rouge1 & Rouge2 \\
\midrule
VAE	&$7.48$ &	$11.09$ &	$31.78$ &	$8.66$ & $2.92$ &	$3.46$ &	$15.13$ &	$3.4$\\
CGMH   & $7.84$   & $11.45$ & $32.19$  & $8.67$   & $4.18$   & $5.32$  & $19.96$  & $5.44$   \\
UPSA   & $9.26$   & $14.16$ & $37.18$  & $11.21$  & $4.93$   & $6.87$  & $28.34$  & $8.53$   \\
PUP    & $\textbf{10.72}$  & $\textbf{15.81}$ & $\textbf{37.38}$  & $\textbf{13.87}$  & $\textbf{6.62}$   & $\textbf{13.03}$ & $\textbf{39.12}$  & $\textbf{12.91}$ \\
\bottomrule
\end{tabular}

\end{table*}

\subsection{Baselines}
\label{competing}
We consider the following unsupervised baselines and domain-adapted approaches for comparison. \newline
\noindent {\bf{UPSA}} is a simulated annealing based approach~\cite{liu2019unsupervised} that attempts to generate paraphrases using a stochastic search algorithm and achieves state-of-art unsupervised paraphrasing results. We use its open source implementation to generate the paraphrases and compare against our approach. \newline
\noindent {\bf{CGMH}} is a Metropolis-Hastings based approach~\cite{miao2019cgmh} that generates paraphrase by constraining the decoder at inference time. We use its open source implementation in our comparisons. 

\noindent {\bf{Domain-adapted models}} are trained in a supervised fashion on one dataset and adapted to another dataset in an unsupervised fashion. For this, we use previously reported results in~\cite{li2019decomposable} for Quora and WikiAnswers datasets. 

We do not compare with the rule-based approaches such as~\cite{mckeown1983paraphrasing,barzilay2003learning} due to the lack of availability of the rules or any implementation. 

\subsection{Evaluation Metrics}
\label{metrics}
We use well-accepted automatic quantitative evaluation metrics as well as qualitative human studies in order to compare the performance of our method against the competing approaches. For quantitative measures, we use BLEU~\cite{papineni2002bleu} and ROUGE~\cite{hovy2006automated} metrics, which have been widely utilized in the previous work to measure the quality of the paraphrases. Additionally, we use i-BLUE~\cite{sun2012joint} by following the metrics in the most recent work~\cite{li2019decomposable,liu2019unsupervised}. The metric i-BLUE~\cite{sun2012joint} aims to measure the diversity of expression in the generated paraphrases by penalizing copying words from input sentences.

\subsection{Implementation Details}
\label{implementation}
The VAE contains two layers with $300$-dimensional LSTM units. Our DRL-based model also has two-layers and uses $300$-dimensional word embeddings (not pre-trained) and $300$-dimensional hidden units.
LSTM is utilized as a recurrent unit, and dropout of $0.5$ is used.
All the sentences are lower cased, and the maximum sentence length is $15$ (i.e., we truncate longer sentences to maintain consistency with previous work). The vocabulary size for each dataset is listed in Table~\ref{tab:datasets}, and infrequent tokens are replaced with $<unk>$ token.
We use Adam optimizer with learning rates of $0.15$, $10^{-3}$, and $10^{-4}$ in the pre-train, transition, and DRL phases, respectively.
The mini-batch size is $32$ and gradient clipping of a maximum gradient norm of $2$ is used in all the phases.
The validation is done after every epoch and the model with the best rewards is saved automatically.
Whether to sample or use argmax, $\kappa=0.9995$ is used. To compute the probability $\delta$, which determines whether to pass VAE's output to the decoder, $l$ is set to $8$ during training.
At inference time, we utilize beam search~\cite{wiseman2016sequence} with a beam size of $b=8$ to sample paraphrases for the given input sentences.
For the reward function, $\alpha =0.4$, $\beta = 0.3$, $\gamma =0.3$, $\tau_{min} = 0.3$, $\tau_{max} = 0.98$, and $\lambda_{min}=0.3$ are used. All the hypterparameters are picked based on the validation split of the Quora dataset, and then consistently used for all the other datasets.

\section{Results}
\label{results}
\begin{table}[t!]
\caption{Subjective human studies on paraphrase generations by unsupervised methods on Quora dataset.}
\label{tab:human}
\begin{tabular}{lrrr}
\toprule
Method & Diversity & Fluency & Similarity \\
\midrule
CGMH   & $3.14$ $\pm$ $0.053$     & $4.1 \pm 0.042$     & $2.97 \pm 0.055$       \\
UPSA   & $2.96 \pm 0.052$     & $4.35 \pm 0.033$    & $3.89 \pm 0.045$       \\
PUP    & $\textbf{3.27} \pm 0.048 $     & $\textbf{4.42} \pm 0.027$    & $\textbf{4.09} \pm 0.035$       \\
\bottomrule       
\end{tabular}
\end{table}

\subsection{Automatic Metrics}
\label{auto}
Table~\ref{tab:quora} presents the performance of unsupervised and domain-adapted methods on the Quora and WikiAnswers datasets; the best method among all is shown in bold and the best among unsupervised methods is underlined for each metric.
Unsupervised methods are trained with non-parallel corpora, and domain-adapted techniques are trained on Quora dataset in a supervised fashion and then domain adapted for WikiAnswers dataset in an unsupervised fashion (and vice versa).  Our proposed method, PUP, outperforms all the unsupervised approaches on all metrics for Quora and WikiAnswers datasets (except Rouge2 for Quora dataset where performance is very competitive with UPSA). Similarly, PUP also outperforms domain-adapted methods for automatic metrics on Quora and WikiAnswers (except i-BLEU for WikiAnswers dataset where the performance is competitive).
Although domain-adapted approaches have the advantage of supervised training on one dataset, this advantage does not transfer effectively to the other dataset despite the similarities between the datasets -- i.e., Quora and WikiAnswers are both questions datasets.
This also highlights that unsupervised approaches are worth exploring for the paraphrasing task as they can be applied to a variety of unlabeled domains or datasets in a flexible way without a need for adaptation.
Moreover, the results for VAE (which we use to pre-train our DRL model) are presented in Table~\ref{tab:quora} and Table~\ref{tab:mscoco} to highlight the performance gain of PUP on each metric. 

Table~\ref{tab:mscoco} presents the results of all unsupervised approaches on MSCOCO and Twitter datasets, where the best model is shown in bold for each metric.
Our proposed method, PUP, is a clear winner on all the metrics among all the unsupervised approaches, which demonstrates the stellar performance of our method as well as the quality of our DRL reward function.
The lower performance of unsupervised methods on Twitter dataset can be ascribed to the noisy tweets data, however, PUP has significantly better performance (i.e., 90\% performance gain on BLEU, and 34\% on i-BLEU scores with respect to UPSA) compared to other methods on all of the metrics, which signifies the robustness of the PUP.

\subsection{Subjective Human Evaluations}
\label{human}
To further illustrate the superior quality of the paraphrases generated by PUP, we conduct subjective human evaluations on Quora dataset. 
Table~\ref{tab:human} presents the average scores along with the confidence intervals of human evaluators for diversity in expression, language fluency, and semantic similarity on randomly selected $300$ paraphrases generated by all three unsupervised methods (CGMH, UPSA, and PUP). We used Amazon Mechanical Turk (a widely-used crowd sourcing platform) in our human studies.
We selected Mechanical Turk \textit{Masters} from the USA with a HIT approval rate of $\geq 90\%$ to rate the paraphrases on a scale of $1-5$ ($1$ being the worst and $5$ the best) for the three evaluation criteria diversity, fluency, and similarity.
Each paraphrase is scored by three different evaluators. Our method PUP outperforms all the competing unsupervised approaches on all criteria. It should also be noted that CGMH is better on diversity of expression than UPSA, and the opposite results are observed for semantic similarity and fluency. In contrast, our reward function facilitates a good balance between the diversity in expression, semantic similarity, and fluency. A similar trend can also be observed in Table~\ref{tab:auto} and Table~\ref{tab:generations}, which present automatically calculated reward  and a few example paraphrases generated by all three unsupervised approaches, respectively.

\begin{table}[t!]
\caption{Performance of the unsupervised methods for the components of the reward function on Quora dataset.}
\label{tab:auto}
\begin{tabular}{lrrrr}
\toprule
Method & Diversity & Fluency & Similarity & Reward \\
\midrule
VAE   & $0.31$      & $0.72$    & $0.47$       & $0.497$  \\
CGMH   & $0.29$      & $0.73$    & $0.49$       & $0.502$  \\
UPSA   & $0.25$      & $0.72$    & $0.68$       & $0.563$  \\
PUP   & $\textbf{0.53}$      & $\textbf{0.95}$    & $\textbf{0.81}$       & $\textbf{0.768}$ \\
\bottomrule 
\end{tabular}
\end{table}

\begin{table*}[t!]
\centering
\footnotesize
\caption{Example paraphrase generations by PUP and other unsupervised competing methods on Quora dataset.}
\label{tab:generations}
\begin{tabular}{rllll}
\toprule
Sr. \# & Input Sentence                                         & CGMH Generation                                       & UPSA Generation                            & PUP Generation                                   \\
\midrule
1. & \begin{tabular}[c]{@{}l@{}}how can i work in microsoft\end{tabular} & \begin{tabular}[c]{@{}l@{}}how can i \underline{prepare for cpt}\end{tabular}                 & \begin{tabular}[c]{@{}l@{}}how can i \underline{get to} work at microsoft\end{tabular}                  & \begin{tabular}[c]{@{}l@{}}how \underline{do i get a job at} microsoft\end{tabular}
\\
2. & \begin{tabular}[c]{@{}l@{}}which is the best shampoo  for \\ dandruff\end{tabular}                 & \begin{tabular}[c]{@{}l@{}}what is the best shampoo  for  \underline{sciatica}\end{tabular}                 & \begin{tabular}[c]{@{}l@{}}which is the best  shampoo for  \underline{oily skin}\end{tabular}    & \begin{tabular}[c]{@{}l@{}}\underline{what are} the \underline{proper shampoos} for \\ dandruff\end{tabular}  \\
3. & \begin{tabular}[c]{@{}l@{}}which book is the best to learn algo\end{tabular}                   & \begin{tabular}[c]{@{}l@{}}which \underline{programming language}  is the \\best to learn algo\end{tabular} & \begin{tabular}[c]{@{}l@{}}which \underline{book is best} to learn algo\end{tabular}           & \begin{tabular}[c]{@{}l@{}}\underline{what} is a best book \underline{for learning}\\ \underline{algos} \end{tabular}           \\
4. &what is the best mac game                                                                         & what is the best \underline{video} game                                                                      & \begin{tabular}[c]{@{}l@{}}what is the best mac \underline{app for android}\\ games\end{tabular} & \begin{tabular}[c]{@{}l@{}}what \underline{are some good} mac \underline{games}\end{tabular}                     \\
5. & what are the reasons of war                                                                       & \begin{tabular}[c]{@{}l@{}}what are the \underline{positive aspects} of\\ nuclear war\end{tabular}          & \begin{tabular}[c]{@{}l@{}}what are the \underline{main reasons for a civil} war\end{tabular}  & \begin{tabular}[c]{@{}l@{}}what is the \underline{main} reason for war\end{tabular}                       
    \\
\bottomrule
\end{tabular}
\end{table*}

\subsection{Evaluation on Reward Function}
\label{reward}

Table~\ref{tab:auto} presents the average scores of all the components of our proposed reward function on Quora dataset for all the unsupervised approaches. Perhaps not surprisingly, our method outperforms other methods on each individual component of the reward by large margin. Intuitively, this arises from the fact that our DRL-based model is explicitly trained to optimize these reward components.
Remarkably, DRL process improves the reward by more than 50\% compared to the pre-training phase, i.e.,~the reward of VAE. This is also visible in Figure~\ref{fig:reward} where PUP starts with a reward value of around $0.5$ and is able to achieve up to $0.77$ towards the end of the last phase of training. 

\subsection{Ablation Study}
\label{ablation}
\begin{figure}[t!]
  \centering
  \includegraphics[width=\linewidth]{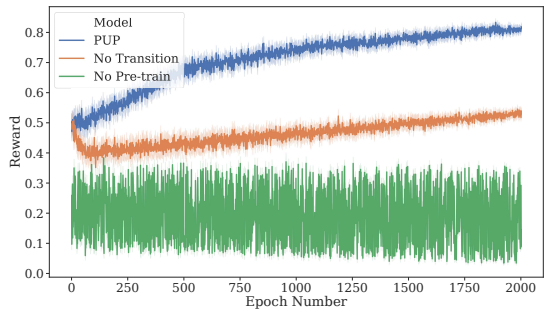}
  \caption{Evolution of the reward value for PUP variants over the course of the training.}
  \label{fig:reward}
\end{figure}

Figure~\ref{fig:reward} presents the rewards achieved over the course of different epochs by three models:
\myNum{i} PUP, pre-trained and uses the transition phase;
\myNum{ii} {\tt No Transition}, pre-trained but does not use the transition phase; and
\myNum{iii} {\tt No Pre-train}, not pre-trained at all.
It highlights the need for the distinct phases in our training procedure.
It can be observed that without the pre-training phase, {\tt No Pre-train} model is unable to maximize the expected reward. The reward remains small and fluctuates randomly. Similarly, transition phase is also required, as abrupt shift from VAE to DRL derails the training for {\tt No Transition} model, whereas PUP is able to rapidly and consistently improve the reward as the number of epochs grow.

\section{Related Work}
\label{related}

The automatic paraphrasing task is one of the common NLP tasks, which has widespread applications. A wide range of approaches were developed to solve this problem. Rule-based~\cite{mckeown1983paraphrasing,ellsworth2007mutaphrase,pavlick2015framenet+,barzilay2003learning} and data-driven approaches~\cite{madnani2010generating,zhao2009application} are some of the earliest techniques. Automatically constructed paraphrase detection datasets using SVM-based classifiers and other unsupervised approaches are introduced in~\cite{Dolan:2004:UCL:1220355.1220406,dolan2005automatically}.

Recently, supervised deep learning approaches have also been used for paraphrase generation. Stacked residual LSTM networks~\cite{prakash2016neural} is one of the earliest efforts in the paraphrase generation utilizing deep networks. \cite{li2017paraphrase} makes use of deep reinforcement learning for paraphrase generation in a supervised fashion. Supervised paraphrase generation using LSTM-based VAE~\cite{gupta2018deep}, transformer model~\cite{vaswani2017attention}, pointer-generator networks~\cite{see2017get} have also shown promising results. Supervised paraphrase generation at different granularity levels (i.e., lexical, phrasal and sentential levels)~\cite{li2019decomposable} is achieved with template learning. Additionally these models can also be adapted to new domains in an unsupervised fashion, utilizing the learned templates with the assumption that both domains share similar templates.

Unsupervised paraphrasing is a challenging and emerging NLP task, and the literature is relatively limited. The VAE~\cite{bowman2015generating} is trained in an unsupervised fashion (i.e., no parallel corpus is required), by maximizing the lower bounds for the log-likelihood. The VAE's decoder can sample sentences (i.e., paraphrases), which are less controllable~\cite{miao2019cgmh}, but serve as a good baseline for the unsupervised paraphrasing task. CGMH~\cite{miao2019cgmh} proposes a constrained sentence generation using Metropolis-Hastings Sampling by adding constraints on the decoder at inference time, and hence does not require parallel corpora. UPSA~\cite{liu2019unsupervised} generates paraphrases by simulated annealing, and achieves state-of-art results on the task. It proposes a search objective function, which involves semantic similarity and fluency for performing diverse word replacement, insertion or deletion operations, thus generating paraphrases in an unsupervised fashion. 
In contrast, we formulate the task as a deep reinforcement learning problem and progressively train the policy to maximize the expected reward, which includes semantic adequacy, language fluency, and diversity in expression. 
\section{Conclusion and Future Work}
\label{conclusion}
We have presented a progressive approach to train a DRL-based unsupervised paraphrasing model.
Our method provides a warm-start to the DRL-based model with a pre-trained VAE (i.e., trained on non-parallel corpus). Then, our model progressively transitions from VAE's output to acting according to its policy.
We also propose a reward function which incorporates all the attributes of a good paraphrase and does not require parallel sentences.
The paraphrases generated by our model outperform both state-of-the-art unsupervised paraphrasing and domain-adapted supervised models on automatic metrics.
Specifically, our method achieves up to 90\% and 34\% performance gains for the BLEU and the i-BLEU  metrics compared to state-of-the-art unsupervised methods, respectively. Moreover, the paraphrases generated by our method were rated the highest by human evaluators for all considered criteria: diversity of expression, fluency, and semantic similarity to input sentences. 
Since our technique is the first to successfully warm-start DRL with an unsupervised model, we plan on investigating the broader implications of our technique on other NLP problems with scarce labeled training data.

\begin{acks}
This work is supported in part by the National Science Foundation (NSF) under grants IIS-1838222, IIS-1901379 and CNS-1932254.
\end{acks}

\bibliographystyle{ACM-Reference-Format}
\bibliography{sample-base}


\begin{thebibliography}{50}


\ifx \showCODEN    \undefined \def \showCODEN     #1{\unskip}     \fi
\ifx \showDOI      \undefined \def \showDOI       #1{#1}\fi
\ifx \showISBNx    \undefined \def \showISBNx     #1{\unskip}     \fi
\ifx \showISBNxiii \undefined \def \showISBNxiii  #1{\unskip}     \fi
\ifx \showISSN     \undefined \def \showISSN      #1{\unskip}     \fi
\ifx \showLCCN     \undefined \def \showLCCN      #1{\unskip}     \fi
\ifx \shownote     \undefined \def \shownote      #1{#1}          \fi
\ifx \showarticletitle \undefined \def \showarticletitle #1{#1}   \fi
\ifx \showURL      \undefined \def \showURL       {\relax}        \fi
\providecommand\bibfield[2]{#2}
\providecommand\bibinfo[2]{#2}
\providecommand\natexlab[1]{#1}
\providecommand\showeprint[2][]{arXiv:#2}

\bibitem[\protect\citeauthoryear{??}{Quo}{[n.d.]}]%
        {QuoraQue91:online}
 \bibinfo{year}{[n.d.]}\natexlab{}.
\newblock \bibinfo{title}{Quora Question Pairs | Kaggle}.
\newblock
  \bibinfo{howpublished}{\url{https://www.kaggle.com/c/quora-question-pairs}}.
\newblock
\newblock
\shownote{(Accessed on 02/14/2020).}


\bibitem[\protect\citeauthoryear{Banerjee and Lavie}{Banerjee and
  Lavie}{2005}]%
        {banerjee2005meteor}
\bibfield{author}{\bibinfo{person}{Satanjeev Banerjee} {and}
  \bibinfo{person}{Alon Lavie}.} \bibinfo{year}{2005}\natexlab{}.
\newblock \showarticletitle{METEOR: An automatic metric for MT evaluation with
  improved correlation with human judgments}. In
  \bibinfo{booktitle}{\emph{Proceedings of the acl workshop on intrinsic and
  extrinsic evaluation measures for machine translation and/or summarization}}.
  \bibinfo{pages}{65--72}.
\newblock


\bibitem[\protect\citeauthoryear{Barzilay and Lee}{Barzilay and Lee}{2003}]%
        {barzilay2003learning}
\bibfield{author}{\bibinfo{person}{Regina Barzilay} {and}
  \bibinfo{person}{Lillian Lee}.} \bibinfo{year}{2003}\natexlab{}.
\newblock \showarticletitle{Learning to paraphrase: an unsupervised approach
  using multiple-sequence alignment}. In \bibinfo{booktitle}{\emph{Proceedings
  of the 2003 Conference of the North American Chapter of the Association for
  Computational Linguistics on Human Language Technology-Volume 1}}.
  Association for Computational Linguistics, \bibinfo{pages}{16--23}.
\newblock


\bibitem[\protect\citeauthoryear{Bengio}{Bengio}{2008}]%
        {bengio2008neural}
\bibfield{author}{\bibinfo{person}{Yoshua Bengio}.}
  \bibinfo{year}{2008}\natexlab{}.
\newblock \showarticletitle{Neural net language models}.
\newblock \bibinfo{journal}{\emph{Scholarpedia}} \bibinfo{volume}{3},
  \bibinfo{number}{1} (\bibinfo{year}{2008}), \bibinfo{pages}{3881}.
\newblock


\bibitem[\protect\citeauthoryear{Bowman, Vilnis, Vinyals, Dai, Jozefowicz, and
  Bengio}{Bowman et~al\mbox{.}}{2015}]%
        {bowman2015generating}
\bibfield{author}{\bibinfo{person}{Samuel~R Bowman}, \bibinfo{person}{Luke
  Vilnis}, \bibinfo{person}{Oriol Vinyals}, \bibinfo{person}{Andrew~M Dai},
  \bibinfo{person}{Rafal Jozefowicz}, {and} \bibinfo{person}{Samy Bengio}.}
  \bibinfo{year}{2015}\natexlab{}.
\newblock \showarticletitle{Generating sentences from a continuous space}.
\newblock \bibinfo{journal}{\emph{arXiv preprint arXiv:1511.06349}}
  (\bibinfo{year}{2015}).
\newblock


\bibitem[\protect\citeauthoryear{Cer, Diab, Agirre, Lopez-Gazpio, and
  Specia}{Cer et~al\mbox{.}}{2017}]%
        {cer2017semeval}
\bibfield{author}{\bibinfo{person}{Daniel Cer}, \bibinfo{person}{Mona Diab},
  \bibinfo{person}{Eneko Agirre}, \bibinfo{person}{Inigo Lopez-Gazpio}, {and}
  \bibinfo{person}{Lucia Specia}.} \bibinfo{year}{2017}\natexlab{}.
\newblock \showarticletitle{Semeval-2017 task 1: Semantic textual
  similarity-multilingual and cross-lingual focused evaluation}.
\newblock \bibinfo{journal}{\emph{arXiv preprint arXiv:1708.00055}}
  (\bibinfo{year}{2017}).
\newblock


\bibitem[\protect\citeauthoryear{Cer, Yang, Kong, Hua, Limtiaco, John,
  Constant, Guajardo-Cespedes, Yuan, Tar, et~al\mbox{.}}{Cer
  et~al\mbox{.}}{2018}]%
        {cer2018universal}
\bibfield{author}{\bibinfo{person}{Daniel Cer}, \bibinfo{person}{Yinfei Yang},
  \bibinfo{person}{Sheng-yi Kong}, \bibinfo{person}{Nan Hua},
  \bibinfo{person}{Nicole Limtiaco}, \bibinfo{person}{Rhomni~St John},
  \bibinfo{person}{Noah Constant}, \bibinfo{person}{Mario Guajardo-Cespedes},
  \bibinfo{person}{Steve Yuan}, \bibinfo{person}{Chris Tar}, {et~al\mbox{.}}}
  \bibinfo{year}{2018}\natexlab{}.
\newblock \showarticletitle{Universal sentence encoder}.
\newblock \bibinfo{journal}{\emph{arXiv preprint arXiv:1803.11175}}
  (\bibinfo{year}{2018}).
\newblock


\bibitem[\protect\citeauthoryear{Chen and Dolan}{Chen and Dolan}{2011}]%
        {chen2011collecting}
\bibfield{author}{\bibinfo{person}{David~L Chen} {and}
  \bibinfo{person}{William~B Dolan}.} \bibinfo{year}{2011}\natexlab{}.
\newblock \showarticletitle{Collecting highly parallel data for paraphrase
  evaluation}. In \bibinfo{booktitle}{\emph{Proceedings of the 49th Annual
  Meeting of the Association for Computational Linguistics: Human Language
  Technologies-Volume 1}}. Association for Computational Linguistics,
  \bibinfo{pages}{190--200}.
\newblock


\bibitem[\protect\citeauthoryear{Daum{\'e}, Langford, and Marcu}{Daum{\'e}
  et~al\mbox{.}}{2009}]%
        {daume2009search}
\bibfield{author}{\bibinfo{person}{Hal Daum{\'e}}, \bibinfo{person}{John
  Langford}, {and} \bibinfo{person}{Daniel Marcu}.}
  \bibinfo{year}{2009}\natexlab{}.
\newblock \showarticletitle{Search-based structured prediction}.
\newblock \bibinfo{journal}{\emph{Machine learning}} \bibinfo{volume}{75},
  \bibinfo{number}{3} (\bibinfo{year}{2009}), \bibinfo{pages}{297--325}.
\newblock


\bibitem[\protect\citeauthoryear{Dayan and Niv}{Dayan and Niv}{2008}]%
        {dayan2008reinforcement}
\bibfield{author}{\bibinfo{person}{Peter Dayan} {and} \bibinfo{person}{Yael
  Niv}.} \bibinfo{year}{2008}\natexlab{}.
\newblock \showarticletitle{Reinforcement learning: the good, the bad and the
  ugly}.
\newblock \bibinfo{journal}{\emph{Current opinion in neurobiology}}
  \bibinfo{volume}{18}, \bibinfo{number}{2} (\bibinfo{year}{2008}),
  \bibinfo{pages}{185--196}.
\newblock


\bibitem[\protect\citeauthoryear{Dolan, Quirk, and Brockett}{Dolan
  et~al\mbox{.}}{2004}]%
        {Dolan:2004:UCL:1220355.1220406}
\bibfield{author}{\bibinfo{person}{Bill Dolan}, \bibinfo{person}{Chris Quirk},
  {and} \bibinfo{person}{Chris Brockett}.} \bibinfo{year}{2004}\natexlab{}.
\newblock \showarticletitle{Unsupervised Construction of Large Paraphrase
  Corpora: Exploiting Massively Parallel News Sources}. In
  \bibinfo{booktitle}{\emph{Proceedings of the 20th International Conference on
  Computational Linguistics}} (Geneva, Switzerland)
  \emph{(\bibinfo{series}{COLING '04})}. \bibinfo{publisher}{Association for
  Computational Linguistics}, \bibinfo{address}{Stroudsburg, PA, USA}, Article
  \bibinfo{articleno}{350}.
\newblock
\urldef\tempurl%
\url{https://doi.org/10.3115/1220355.1220406}
\showDOI{\tempurl}


\bibitem[\protect\citeauthoryear{Dolan and Brockett}{Dolan and
  Brockett}{2005}]%
        {dolan2005automatically}
\bibfield{author}{\bibinfo{person}{William~B Dolan} {and}
  \bibinfo{person}{Chris Brockett}.} \bibinfo{year}{2005}\natexlab{}.
\newblock \showarticletitle{Automatically constructing a corpus of sentential
  paraphrases}. In \bibinfo{booktitle}{\emph{Proceedings of the Third
  International Workshop on Paraphrasing (IWP2005)}}.
\newblock


\bibitem[\protect\citeauthoryear{Ellsworth and Janin}{Ellsworth and
  Janin}{2007}]%
        {ellsworth2007mutaphrase}
\bibfield{author}{\bibinfo{person}{Michael Ellsworth} {and}
  \bibinfo{person}{Adam Janin}.} \bibinfo{year}{2007}\natexlab{}.
\newblock \showarticletitle{Mutaphrase: Paraphrasing with framenet}. In
  \bibinfo{booktitle}{\emph{Proceedings of the ACL-PASCAL Workshop on Textual
  Entailment and Paraphrasing}}. Association for Computational Linguistics,
  \bibinfo{pages}{143--150}.
\newblock


\bibitem[\protect\citeauthoryear{Fader, Zettlemoyer, and Etzioni}{Fader
  et~al\mbox{.}}{2013}]%
        {fader2013paraphrase}
\bibfield{author}{\bibinfo{person}{Anthony Fader}, \bibinfo{person}{Luke
  Zettlemoyer}, {and} \bibinfo{person}{Oren Etzioni}.}
  \bibinfo{year}{2013}\natexlab{}.
\newblock \showarticletitle{Paraphrase-driven learning for open question
  answering}. In \bibinfo{booktitle}{\emph{Proceedings of the 51st Annual
  Meeting of the Association for Computational Linguistics (Volume 1: Long
  Papers)}}. \bibinfo{pages}{1608--1618}.
\newblock


\bibitem[\protect\citeauthoryear{Gupta, Agarwal, Singh, and Rai}{Gupta
  et~al\mbox{.}}{2018}]%
        {gupta2018deep}
\bibfield{author}{\bibinfo{person}{Ankush Gupta}, \bibinfo{person}{Arvind
  Agarwal}, \bibinfo{person}{Prawaan Singh}, {and} \bibinfo{person}{Piyush
  Rai}.} \bibinfo{year}{2018}\natexlab{}.
\newblock \showarticletitle{A deep generative framework for paraphrase
  generation}. In \bibinfo{booktitle}{\emph{Thirty-Second AAAI Conference on
  Artificial Intelligence}}.
\newblock


\bibitem[\protect\citeauthoryear{Heafield}{Heafield}{2011}]%
        {heafield2011kenlm}
\bibfield{author}{\bibinfo{person}{Kenneth Heafield}.}
  \bibinfo{year}{2011}\natexlab{}.
\newblock \showarticletitle{KenLM: Faster and smaller language model queries}.
  In \bibinfo{booktitle}{\emph{Proceedings of the sixth workshop on statistical
  machine translation}}. Association for Computational Linguistics,
  \bibinfo{pages}{187--197}.
\newblock


\bibitem[\protect\citeauthoryear{Hochreiter and Schmidhuber}{Hochreiter and
  Schmidhuber}{1997}]%
        {hochreiter1997lstm}
\bibfield{author}{\bibinfo{person}{Sepp Hochreiter} {and}
  \bibinfo{person}{J{\"u}rgen Schmidhuber}.} \bibinfo{year}{1997}\natexlab{}.
\newblock \showarticletitle{LSTM can solve hard long time lag problems}. In
  \bibinfo{booktitle}{\emph{Advances in neural information processing
  systems}}. \bibinfo{pages}{473--479}.
\newblock


\bibitem[\protect\citeauthoryear{Hovy, Lin, Zhou, and Fukumoto}{Hovy
  et~al\mbox{.}}{2006}]%
        {hovy2006automated}
\bibfield{author}{\bibinfo{person}{Eduard~H Hovy}, \bibinfo{person}{Chin-Yew
  Lin}, \bibinfo{person}{Liang Zhou}, {and} \bibinfo{person}{Junichi
  Fukumoto}.} \bibinfo{year}{2006}\natexlab{}.
\newblock \showarticletitle{Automated Summarization Evaluation with Basic
  Elements.}. In \bibinfo{booktitle}{\emph{LREC}}, Vol.~\bibinfo{volume}{6}.
  Citeseer, \bibinfo{pages}{899--902}.
\newblock


\bibitem[\protect\citeauthoryear{Kingma and Welling}{Kingma and
  Welling}{2013}]%
        {kingma2013auto}
\bibfield{author}{\bibinfo{person}{Diederik~P Kingma} {and}
  \bibinfo{person}{Max Welling}.} \bibinfo{year}{2013}\natexlab{}.
\newblock \showarticletitle{Auto-encoding variational bayes}.
\newblock \bibinfo{journal}{\emph{arXiv preprint arXiv:1312.6114}}
  (\bibinfo{year}{2013}).
\newblock


\bibitem[\protect\citeauthoryear{Kissner}{Kissner}{2006}]%
        {kissner2006summarizing}
\bibfield{author}{\bibinfo{person}{Emily Kissner}.}
  \bibinfo{year}{2006}\natexlab{}.
\newblock \showarticletitle{Summarizing, paraphrasing and retelling}.
\newblock \bibinfo{journal}{\emph{Portsmouth, NH: Heinernann}}
  (\bibinfo{year}{2006}).
\newblock


\bibitem[\protect\citeauthoryear{Knight and Marcu}{Knight and Marcu}{2000}]%
        {knight2000statistics}
\bibfield{author}{\bibinfo{person}{Kevin Knight} {and} \bibinfo{person}{Daniel
  Marcu}.} \bibinfo{year}{2000}\natexlab{}.
\newblock \showarticletitle{Statistics-based summarization-step one: Sentence
  compression}.
\newblock \bibinfo{journal}{\emph{AAAI/IAAI}}  \bibinfo{volume}{2000}
  (\bibinfo{year}{2000}), \bibinfo{pages}{703--710}.
\newblock


\bibitem[\protect\citeauthoryear{Lan, Qiu, He, and Xu}{Lan
  et~al\mbox{.}}{2017}]%
        {lan2017continuously}
\bibfield{author}{\bibinfo{person}{Wuwei Lan}, \bibinfo{person}{Siyu Qiu},
  \bibinfo{person}{Hua He}, {and} \bibinfo{person}{Wei Xu}.}
  \bibinfo{year}{2017}\natexlab{}.
\newblock \showarticletitle{A continuously growing dataset of sentential
  paraphrases}.
\newblock \bibinfo{journal}{\emph{arXiv preprint arXiv:1708.00391}}
  (\bibinfo{year}{2017}).
\newblock


\bibitem[\protect\citeauthoryear{Li, Jiang, Shang, and Li}{Li
  et~al\mbox{.}}{2017}]%
        {li2017paraphrase}
\bibfield{author}{\bibinfo{person}{Zichao Li}, \bibinfo{person}{Xin Jiang},
  \bibinfo{person}{Lifeng Shang}, {and} \bibinfo{person}{Hang Li}.}
  \bibinfo{year}{2017}\natexlab{}.
\newblock \showarticletitle{Paraphrase generation with deep reinforcement
  learning}.
\newblock \bibinfo{journal}{\emph{arXiv preprint arXiv:1711.00279}}
  (\bibinfo{year}{2017}).
\newblock


\bibitem[\protect\citeauthoryear{Li, Jiang, Shang, and Liu}{Li
  et~al\mbox{.}}{2019}]%
        {li2019decomposable}
\bibfield{author}{\bibinfo{person}{Zichao Li}, \bibinfo{person}{Xin Jiang},
  \bibinfo{person}{Lifeng Shang}, {and} \bibinfo{person}{Qun Liu}.}
  \bibinfo{year}{2019}\natexlab{}.
\newblock \showarticletitle{Decomposable neural paraphrase generation}.
\newblock \bibinfo{journal}{\emph{arXiv preprint arXiv:1906.09741}}
  (\bibinfo{year}{2019}).
\newblock


\bibitem[\protect\citeauthoryear{Lin, Maire, Belongie, Hays, Perona, Ramanan,
  Doll{\'a}r, and Zitnick}{Lin et~al\mbox{.}}{2014}]%
        {lin2014microsoft}
\bibfield{author}{\bibinfo{person}{Tsung-Yi Lin}, \bibinfo{person}{Michael
  Maire}, \bibinfo{person}{Serge Belongie}, \bibinfo{person}{James Hays},
  \bibinfo{person}{Pietro Perona}, \bibinfo{person}{Deva Ramanan},
  \bibinfo{person}{Piotr Doll{\'a}r}, {and} \bibinfo{person}{C~Lawrence
  Zitnick}.} \bibinfo{year}{2014}\natexlab{}.
\newblock \showarticletitle{Microsoft coco: Common objects in context}. In
  \bibinfo{booktitle}{\emph{European conference on computer vision}}. Springer,
  \bibinfo{pages}{740--755}.
\newblock


\bibitem[\protect\citeauthoryear{Liu, Mou, Meng, Zhou, Zhou, and Song}{Liu
  et~al\mbox{.}}{2019}]%
        {liu2019unsupervised}
\bibfield{author}{\bibinfo{person}{Xianggen Liu}, \bibinfo{person}{Lili Mou},
  \bibinfo{person}{Fandong Meng}, \bibinfo{person}{Hao Zhou},
  \bibinfo{person}{Jie Zhou}, {and} \bibinfo{person}{Sen Song}.}
  \bibinfo{year}{2019}\natexlab{}.
\newblock \showarticletitle{Unsupervised Paraphrasing by Simulated Annealing}.
\newblock \bibinfo{journal}{\emph{arXiv preprint arXiv:1909.03588}}
  (\bibinfo{year}{2019}).
\newblock


\bibitem[\protect\citeauthoryear{Madnani and Dorr}{Madnani and Dorr}{2010}]%
        {madnani2010generating}
\bibfield{author}{\bibinfo{person}{Nitin Madnani} {and}
  \bibinfo{person}{Bonnie~J Dorr}.} \bibinfo{year}{2010}\natexlab{}.
\newblock \showarticletitle{Generating phrasal and sentential paraphrases: A
  survey of data-driven methods}.
\newblock \bibinfo{journal}{\emph{Computational Linguistics}}
  \bibinfo{volume}{36}, \bibinfo{number}{3} (\bibinfo{year}{2010}),
  \bibinfo{pages}{341--387}.
\newblock


\bibitem[\protect\citeauthoryear{McKeown}{McKeown}{1983}]%
        {mckeown1983paraphrasing}
\bibfield{author}{\bibinfo{person}{Kathleen~R McKeown}.}
  \bibinfo{year}{1983}\natexlab{}.
\newblock \showarticletitle{Paraphrasing questions using given and new
  information}.
\newblock \bibinfo{journal}{\emph{Computational Linguistics}}
  \bibinfo{volume}{9}, \bibinfo{number}{1} (\bibinfo{year}{1983}),
  \bibinfo{pages}{1--10}.
\newblock


\bibitem[\protect\citeauthoryear{Metzler, Hovy, and Zhang}{Metzler
  et~al\mbox{.}}{2011}]%
        {metzler2011empirical}
\bibfield{author}{\bibinfo{person}{Donald Metzler}, \bibinfo{person}{Eduard
  Hovy}, {and} \bibinfo{person}{Chunliang Zhang}.}
  \bibinfo{year}{2011}\natexlab{}.
\newblock \showarticletitle{An empirical evaluation of data-driven paraphrase
  generation techniques}. In \bibinfo{booktitle}{\emph{Proceedings of the 49th
  Annual Meeting of the Association for Computational Linguistics: Human
  Language Technologies: short papers-Volume 2}}. Association for Computational
  Linguistics, \bibinfo{pages}{546--551}.
\newblock


\bibitem[\protect\citeauthoryear{Miao, Zhou, Mou, Yan, and Li}{Miao
  et~al\mbox{.}}{2019}]%
        {miao2019cgmh}
\bibfield{author}{\bibinfo{person}{Ning Miao}, \bibinfo{person}{Hao Zhou},
  \bibinfo{person}{Lili Mou}, \bibinfo{person}{Rui Yan}, {and}
  \bibinfo{person}{Lei Li}.} \bibinfo{year}{2019}\natexlab{}.
\newblock \showarticletitle{Cgmh: Constrained sentence generation by
  metropolis-hastings sampling}. In \bibinfo{booktitle}{\emph{Proceedings of
  the AAAI Conference on Artificial Intelligence}}, Vol.~\bibinfo{volume}{33}.
  \bibinfo{pages}{6834--6842}.
\newblock


\bibitem[\protect\citeauthoryear{Mnih, Kavukcuoglu, Silver, Graves, Antonoglou,
  Wierstra, and Riedmiller}{Mnih et~al\mbox{.}}{2013}]%
        {mnih2013playing}
\bibfield{author}{\bibinfo{person}{Volodymyr Mnih}, \bibinfo{person}{Koray
  Kavukcuoglu}, \bibinfo{person}{David Silver}, \bibinfo{person}{Alex Graves},
  \bibinfo{person}{Ioannis Antonoglou}, \bibinfo{person}{Daan Wierstra}, {and}
  \bibinfo{person}{Martin Riedmiller}.} \bibinfo{year}{2013}\natexlab{}.
\newblock \showarticletitle{Playing atari with deep reinforcement learning}.
\newblock \bibinfo{journal}{\emph{arXiv preprint arXiv:1312.5602}}
  (\bibinfo{year}{2013}).
\newblock


\bibitem[\protect\citeauthoryear{Papineni, Roukos, Ward, and Zhu}{Papineni
  et~al\mbox{.}}{2002}]%
        {papineni2002bleu}
\bibfield{author}{\bibinfo{person}{Kishore Papineni}, \bibinfo{person}{Salim
  Roukos}, \bibinfo{person}{Todd Ward}, {and} \bibinfo{person}{Wei-Jing Zhu}.}
  \bibinfo{year}{2002}\natexlab{}.
\newblock \showarticletitle{BLEU: a method for automatic evaluation of machine
  translation}. In \bibinfo{booktitle}{\emph{Proceedings of the 40th annual
  meeting on association for computational linguistics}}. Association for
  Computational Linguistics, \bibinfo{pages}{311--318}.
\newblock


\bibitem[\protect\citeauthoryear{Pavlick, Wolfe, Rastogi, Callison-Burch,
  Dredze, and Van~Durme}{Pavlick et~al\mbox{.}}{2015}]%
        {pavlick2015framenet+}
\bibfield{author}{\bibinfo{person}{Ellie Pavlick}, \bibinfo{person}{Travis
  Wolfe}, \bibinfo{person}{Pushpendre Rastogi}, \bibinfo{person}{Chris
  Callison-Burch}, \bibinfo{person}{Mark Dredze}, {and}
  \bibinfo{person}{Benjamin Van~Durme}.} \bibinfo{year}{2015}\natexlab{}.
\newblock \showarticletitle{Framenet+: Fast paraphrastic tripling of framenet}.
  In \bibinfo{booktitle}{\emph{Proceedings of the 53rd Annual Meeting of the
  Association for Computational Linguistics and the 7th International Joint
  Conference on Natural Language Processing (Volume 2: Short Papers)}}.
  \bibinfo{pages}{408--413}.
\newblock


\bibitem[\protect\citeauthoryear{Prakash, Hasan, Lee, Datla, Qadir, Liu, and
  Farri}{Prakash et~al\mbox{.}}{2016}]%
        {prakash2016neural}
\bibfield{author}{\bibinfo{person}{Aaditya Prakash}, \bibinfo{person}{Sadid~A
  Hasan}, \bibinfo{person}{Kathy Lee}, \bibinfo{person}{Vivek Datla},
  \bibinfo{person}{Ashequl Qadir}, \bibinfo{person}{Joey Liu}, {and}
  \bibinfo{person}{Oladimeji Farri}.} \bibinfo{year}{2016}\natexlab{}.
\newblock \showarticletitle{Neural paraphrase generation with stacked residual
  LSTM networks}.
\newblock \bibinfo{journal}{\emph{arXiv preprint arXiv:1610.03098}}
  (\bibinfo{year}{2016}).
\newblock


\bibitem[\protect\citeauthoryear{Rezende, Mohamed, and Wierstra}{Rezende
  et~al\mbox{.}}{2014}]%
        {rezende2014stochastic}
\bibfield{author}{\bibinfo{person}{Danilo~Jimenez Rezende},
  \bibinfo{person}{Shakir Mohamed}, {and} \bibinfo{person}{Daan Wierstra}.}
  \bibinfo{year}{2014}\natexlab{}.
\newblock \showarticletitle{Stochastic backpropagation and approximate
  inference in deep generative models}.
\newblock \bibinfo{journal}{\emph{arXiv preprint arXiv:1401.4082}}
  (\bibinfo{year}{2014}).
\newblock


\bibitem[\protect\citeauthoryear{Ross, Gordon, and Bagnell}{Ross
  et~al\mbox{.}}{2011}]%
        {ross2011reduction}
\bibfield{author}{\bibinfo{person}{St{\'e}phane Ross},
  \bibinfo{person}{Geoffrey Gordon}, {and} \bibinfo{person}{Drew Bagnell}.}
  \bibinfo{year}{2011}\natexlab{}.
\newblock \showarticletitle{A reduction of imitation learning and structured
  prediction to no-regret online learning}. In
  \bibinfo{booktitle}{\emph{Proceedings of the fourteenth international
  conference on artificial intelligence and statistics}}.
  \bibinfo{pages}{627--635}.
\newblock


\bibitem[\protect\citeauthoryear{See, Liu, and Manning}{See
  et~al\mbox{.}}{2017}]%
        {see2017get}
\bibfield{author}{\bibinfo{person}{Abigail See}, \bibinfo{person}{Peter~J Liu},
  {and} \bibinfo{person}{Christopher~D Manning}.}
  \bibinfo{year}{2017}\natexlab{}.
\newblock \showarticletitle{Get to the point: Summarization with
  pointer-generator networks}.
\newblock \bibinfo{journal}{\emph{arXiv preprint arXiv:1704.04368}}
  (\bibinfo{year}{2017}).
\newblock


\bibitem[\protect\citeauthoryear{Shah, Hakkani-T{\"u}r, T{\"u}r, Rastogi,
  Bapna, Nayak, and Heck}{Shah et~al\mbox{.}}{2018}]%
        {shah2018building}
\bibfield{author}{\bibinfo{person}{Pararth Shah}, \bibinfo{person}{Dilek
  Hakkani-T{\"u}r}, \bibinfo{person}{Gokhan T{\"u}r}, \bibinfo{person}{Abhinav
  Rastogi}, \bibinfo{person}{Ankur Bapna}, \bibinfo{person}{Neha Nayak}, {and}
  \bibinfo{person}{Larry Heck}.} \bibinfo{year}{2018}\natexlab{}.
\newblock \showarticletitle{Building a conversational agent overnight with
  dialogue self-play}.
\newblock \bibinfo{journal}{\emph{arXiv preprint arXiv:1801.04871}}
  (\bibinfo{year}{2018}).
\newblock


\bibitem[\protect\citeauthoryear{Silver, Schrittwieser, Simonyan, Antonoglou,
  Huang, Guez, Hubert, Baker, Lai, Bolton, et~al\mbox{.}}{Silver
  et~al\mbox{.}}{2017}]%
        {silver2017mastering}
\bibfield{author}{\bibinfo{person}{David Silver}, \bibinfo{person}{Julian
  Schrittwieser}, \bibinfo{person}{Karen Simonyan}, \bibinfo{person}{Ioannis
  Antonoglou}, \bibinfo{person}{Aja Huang}, \bibinfo{person}{Arthur Guez},
  \bibinfo{person}{Thomas Hubert}, \bibinfo{person}{Lucas Baker},
  \bibinfo{person}{Matthew Lai}, \bibinfo{person}{Adrian Bolton},
  {et~al\mbox{.}}} \bibinfo{year}{2017}\natexlab{}.
\newblock \showarticletitle{Mastering the game of go without human knowledge}.
\newblock \bibinfo{journal}{\emph{Nature}} \bibinfo{volume}{550},
  \bibinfo{number}{7676} (\bibinfo{year}{2017}), \bibinfo{pages}{354}.
\newblock


\bibitem[\protect\citeauthoryear{Snover, Dorr, Schwartz, Micciulla, and
  Makhoul}{Snover et~al\mbox{.}}{2006}]%
        {snover2006study}
\bibfield{author}{\bibinfo{person}{Matthew Snover}, \bibinfo{person}{Bonnie
  Dorr}, \bibinfo{person}{Richard Schwartz}, \bibinfo{person}{Linnea
  Micciulla}, {and} \bibinfo{person}{John Makhoul}.}
  \bibinfo{year}{2006}\natexlab{}.
\newblock \showarticletitle{A study of translation edit rate with targeted
  human annotation}. In \bibinfo{booktitle}{\emph{Proceedings of association
  for machine translation in the Americas}}, Vol.~\bibinfo{volume}{200}.
\newblock


\bibitem[\protect\citeauthoryear{Sun and Zhou}{Sun and Zhou}{2012}]%
        {sun2012joint}
\bibfield{author}{\bibinfo{person}{Hong Sun} {and} \bibinfo{person}{Ming
  Zhou}.} \bibinfo{year}{2012}\natexlab{}.
\newblock \showarticletitle{Joint learning of a dual SMT system for paraphrase
  generation}. In \bibinfo{booktitle}{\emph{Proceedings of the 50th Annual
  Meeting of the Association for Computational Linguistics: Short Papers-Volume
  2}}. Association for Computational Linguistics, \bibinfo{pages}{38--42}.
\newblock


\bibitem[\protect\citeauthoryear{Sutskever, Vinyals, and Le}{Sutskever
  et~al\mbox{.}}{2014}]%
        {sutskever2014sequence}
\bibfield{author}{\bibinfo{person}{Ilya Sutskever}, \bibinfo{person}{Oriol
  Vinyals}, {and} \bibinfo{person}{Quoc~V Le}.}
  \bibinfo{year}{2014}\natexlab{}.
\newblock \showarticletitle{Sequence to sequence learning with neural
  networks}. In \bibinfo{booktitle}{\emph{Advances in neural information
  processing systems}}. \bibinfo{pages}{3104--3112}.
\newblock


\bibitem[\protect\citeauthoryear{Sutton, Barto, et~al\mbox{.}}{Sutton
  et~al\mbox{.}}{1998}]%
        {sutton1998introduction}
\bibfield{author}{\bibinfo{person}{Richard~S Sutton}, \bibinfo{person}{Andrew~G
  Barto}, {et~al\mbox{.}}} \bibinfo{year}{1998}\natexlab{}.
\newblock \bibinfo{booktitle}{\emph{Introduction to reinforcement learning}}.
  Vol.~\bibinfo{volume}{135}.
\newblock \bibinfo{publisher}{MIT press Cambridge}.
\newblock


\bibitem[\protect\citeauthoryear{Vaswani, Shazeer, Parmar, Uszkoreit, Jones,
  Gomez, Kaiser, and Polosukhin}{Vaswani et~al\mbox{.}}{2017}]%
        {vaswani2017attention}
\bibfield{author}{\bibinfo{person}{Ashish Vaswani}, \bibinfo{person}{Noam
  Shazeer}, \bibinfo{person}{Niki Parmar}, \bibinfo{person}{Jakob Uszkoreit},
  \bibinfo{person}{Llion Jones}, \bibinfo{person}{Aidan~N Gomez},
  \bibinfo{person}{{\L}ukasz Kaiser}, {and} \bibinfo{person}{Illia
  Polosukhin}.} \bibinfo{year}{2017}\natexlab{}.
\newblock \showarticletitle{Attention is all you need}. In
  \bibinfo{booktitle}{\emph{Advances in neural information processing
  systems}}. \bibinfo{pages}{5998--6008}.
\newblock


\bibitem[\protect\citeauthoryear{Warstadt, Singh, and Bowman}{Warstadt
  et~al\mbox{.}}{2019}]%
        {warstadt2019neural}
\bibfield{author}{\bibinfo{person}{Alex Warstadt}, \bibinfo{person}{Amanpreet
  Singh}, {and} \bibinfo{person}{Samuel~R Bowman}.}
  \bibinfo{year}{2019}\natexlab{}.
\newblock \showarticletitle{Neural network acceptability judgments}.
\newblock \bibinfo{journal}{\emph{Transactions of the Association for
  Computational Linguistics}}  \bibinfo{volume}{7} (\bibinfo{year}{2019}),
  \bibinfo{pages}{625--641}.
\newblock


\bibitem[\protect\citeauthoryear{Williams}{Williams}{1992}]%
        {williams1992simple}
\bibfield{author}{\bibinfo{person}{Ronald~J Williams}.}
  \bibinfo{year}{1992}\natexlab{}.
\newblock \showarticletitle{Simple statistical gradient-following algorithms
  for connectionist reinforcement learning}.
\newblock \bibinfo{journal}{\emph{Machine learning}} \bibinfo{volume}{8},
  \bibinfo{number}{3-4} (\bibinfo{year}{1992}), \bibinfo{pages}{229--256}.
\newblock


\bibitem[\protect\citeauthoryear{Wiseman and Rush}{Wiseman and Rush}{2016}]%
        {wiseman2016sequence}
\bibfield{author}{\bibinfo{person}{Sam Wiseman} {and}
  \bibinfo{person}{Alexander~M Rush}.} \bibinfo{year}{2016}\natexlab{}.
\newblock \showarticletitle{Sequence-to-sequence learning as beam-search
  optimization}.
\newblock \bibinfo{journal}{\emph{arXiv preprint arXiv:1606.02960}}
  (\bibinfo{year}{2016}).
\newblock


\bibitem[\protect\citeauthoryear{Zhao, Lan, Liu, and Li}{Zhao
  et~al\mbox{.}}{2009}]%
        {zhao2009application}
\bibfield{author}{\bibinfo{person}{Shiqi Zhao}, \bibinfo{person}{Xiang Lan},
  \bibinfo{person}{Ting Liu}, {and} \bibinfo{person}{Sheng Li}.}
  \bibinfo{year}{2009}\natexlab{}.
\newblock \showarticletitle{Application-driven statistical paraphrase
  generation}. In \bibinfo{booktitle}{\emph{Proceedings of the Joint Conference
  of the 47th Annual Meeting of the ACL and the 4th International Joint
  Conference on Natural Language Processing of the AFNLP: Volume 2-Volume 2}}.
  Association for Computational Linguistics, \bibinfo{pages}{834--842}.
\newblock


\bibitem[\protect\citeauthoryear{Zhao and Wang}{Zhao and Wang}{2010}]%
        {zhao2010paraphrases}
\bibfield{author}{\bibinfo{person}{Shiqi Zhao} {and} \bibinfo{person}{Haifeng
  Wang}.} \bibinfo{year}{2010}\natexlab{}.
\newblock \showarticletitle{Paraphrases and applications}. In
  \bibinfo{booktitle}{\emph{Coling 2010: Paraphrases and Applications--Tutorial
  notes}}. \bibinfo{pages}{1--87}.
\newblock


\bibitem[\protect\citeauthoryear{Zhao, Wang, Lan, and Liu}{Zhao
  et~al\mbox{.}}{2010}]%
        {zhao2010leveraging}
\bibfield{author}{\bibinfo{person}{Shiqi Zhao}, \bibinfo{person}{Haifeng Wang},
  \bibinfo{person}{Xiang Lan}, {and} \bibinfo{person}{Ting Liu}.}
  \bibinfo{year}{2010}\natexlab{}.
\newblock \showarticletitle{Leveraging multiple MT engines for paraphrase
  generation}. In \bibinfo{booktitle}{\emph{Proceedings of the 23rd
  International Conference on Computational Linguistics}}. Association for
  Computational Linguistics, \bibinfo{pages}{1326--1334}.
\newblock


\end{thebibliography}

\appendix

\section{Supplementary Material}
\balance

\subsection{Human Evaluations Details}
\label{supp:human-eval}
A set of $300$ randomly selected sentences from the test test of the Quora dataset were used for evaluation by crowd workers. The paraphrases generated by every model (i.e., CGMH, UPSA, PUP) were rated by three different crowd workers on the following criteria:
\begin{itemize}
  \item \textbf{Semantic Similarity}: how close is the meaning of paraphrased sentence to the original sentence (i.e., $5$ means same meaning, and 1 means completely different meaning).
  \item \textbf{Fluency}: whether the paraphrased sentence is grammatically acceptable (i.e., $5$ means grammatically correct and $1$ means that it makes no sense).
  \item \textbf{Diversity in expression}: whether different words are used in the paraphrased sentence with respect to the original sentence (i.e., $5$ means at least half of the words are new, and $1$ means that it makes no changes other than stop-words).  
  \end{itemize}

The raters were also provided with the positive (i.e., good example for each criteria) and negative  (i.e., poor example for each criteria) examples for a sample sentence.

\noindent \textbf{Test Sentence:} To avoid carelessly filled responses, a test sentence (negative example) was placed with the three paraphrase generations (one from each model) for each input sentence, which was used to discard the rating provided by that particular worker for the paraphrases of that sentence. The workers were informed about the test sentence in the instructions. The responses of the workers who rated the sentence~>~2  were discarded from the further analysis, which is reported in Section~\ref{human}. However, workers were still paid.

There were a total of three test sentences; one of these was randomly placed in each set (three paraphrases by model, and one test sentence). The test sentence was easy to spot for: 1) totally different meaning than input (i.e., should get $1$ on semantic similarity), 2) totally wrong for grammar correctness (i.e., should get $1$ on fluency), and 3) same copy of the input (i.e., should get $1$ on diversity in expression). 

\subsection{Datasets Preprocessing}
\label{preprocessing}
We perform some of the standard pre-processing steps on all the datasets, which are briefly explained in the main paper as well. In this section, we explain the exact pre-processing steps. We use spaCy
to tokenize the sentences. The maximum sentence length is set to $15$, and longer sentences are trimmed (i.e., to remain consistent with previous works and easy comparison). We further pre-process and build vocabulary using torchtext
by setting \textit{init\_token} (i.e., start of sentence) to $<$sos$>$, \textit{eos\_token} (i.e., end of sentence) to $<$eos$>$, and \textit{lower} (i.e., lower case) to \textit{True}. We also set \textit{min\_freq} (i.e., minimum frequency) to $4$, \textit{unk\_init} (i.e.; infrequent/unknown token replacement) to $<$unk$>$ for all the datasets, and we set \textit{max\_size} (i.e., vocabulary size) to $8$K, $8$K, $10$K, and $8$K for Quora, WikiAnswers, MSCOCO, and Twitter datasets respectively. No pre-trained word embeddings are utilized, instead embeddings are trained while models are being trained. Both the VAE, and DRL model share the same vocabulary.

\subsection{Training Details}
\label{supp:training}
All the hyperparameters are described in Section~\ref{implementation}. We follow the following steps to train the model for each dataset:

\begin{itemize}
  \item The dataset is preprocessed as explained in Section~\ref{preprocessing}. All the datasets used for the experiments are publicly available.
  \item Variational Autoencoder (VAE) is trained for $15$ epochs, which provides a warm-start to the our deep reinforcement learning-based model.
  \item The deep reinforcement learning-based unsupervised paraphrasing model is pre-trained for $15$ epochs with the VAE.
  \item Then the model is trained in the transition, and DRL phases for $2000$ epochs with the same parameters, as explained in Section~\ref{implementation}.
  \item The weights for each component of the reward function, and the values for the thresholds are given in Section~\ref{implementation}.
  \item The model that achieves best reward on the validation set is stored to generate paraphrases on the test-test for automatic evaluation metrics and human studies.
\end{itemize}


\end{document}